# Online Learning of Correspondences between Images

Michael Felsberg, *Member, IEEE,* Fredrik Larsson, Johan Wiklund, Niclas Wadströmer, and Jörgen Ahlberg

**Abstract**—We propose a novel method for iterative learning of point correspondences between image sequences. Points moving on surfaces in 3D space are projected into two images. Given a point in either view, the considered problem is to determine the corresponding location in the other view. The geometry and distortions of the projections are unknown as is the shape of the surface. Given several pairs of point-sets but no access to the 3D scene, correspondence mappings can be found by excessive global optimization or by the fundamental matrix if a perspective projective model is assumed. However, an *iterative* solution on sequences of point-set pairs with *general* imaging geometry is preferable. We derive such a method that optimizes the mapping based on Neyman's chi-square divergence between the densities representing the uncertainties of the estimated and the actual locations. The densities are represented as channel vectors computed with a basis function approach. The mapping between these vectors is updated with each new pair of images such that fast convergence and high accuracy are achieved. The resulting algorithm runs in real-time and is superior to state-of-the-art methods in terms of convergence and accuracy in a number of experiments.

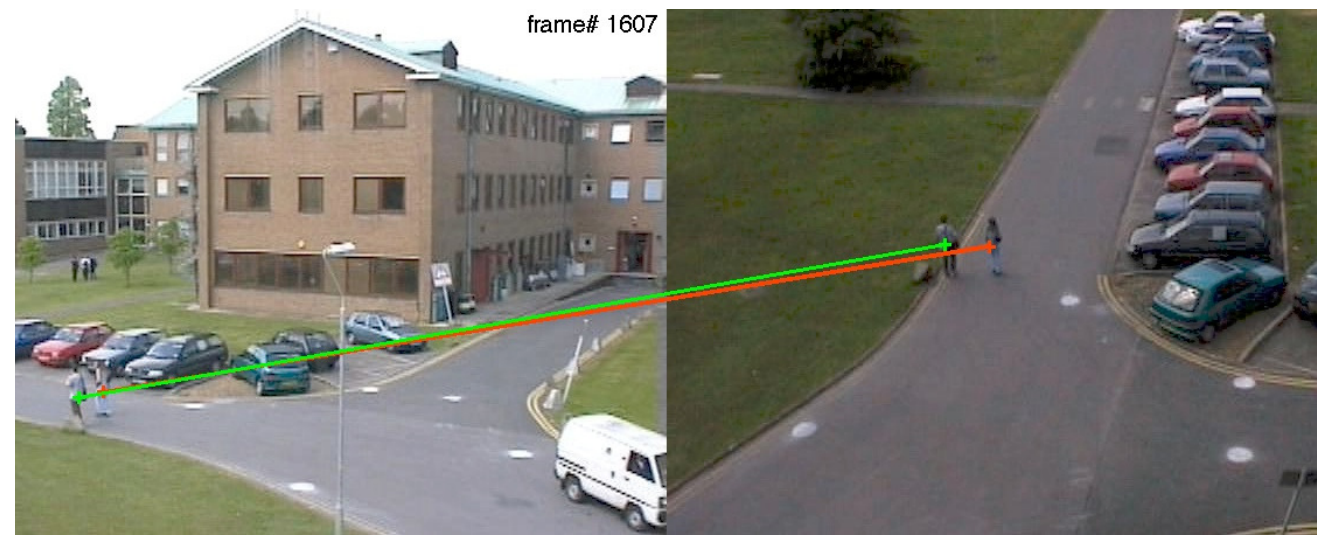

Fig. 1. Example for CMap in a surveillance application.

## 1 Introduction

In multi-view computer vision, a dynamic 3D scene is projected into several images. If the 3D points move (or lie) on 2D surfaces, such as discontinuous curved surfaces, the resulting views show the same scene under multi-modal piecewise continuous distortions. In order to fuse information from the separate views, the corresponding image points of the 3D points are required.

In the two camera case, a point correspondence is defined as the pair of points from the two images that correspond to the same 3D point. The mapping from a point in one image to its corresponding point is potentially multimodal (*e.g.* occlusion, transparency) and spatially uncertain, since points are extracted using error-prone feature detectors. Therefore, points are represented by a probability hypothesis density (PHD) [1] that models the spatial distribution of an a-priori unknown number of points. In contrast to the single point case modeled by a probability density, a PHD combines weighted densities of many hypotheses. Here, PHDs are represented using *channels* [2]. The *correspondence mapping* (CMap) maps the PHD of one image to the PHD of the other image and contains implicitly the geometry of the 2D surfaces and the geometry and distortions of both views.

The CMap is useful in many applications. In video surveillance, see Figure 1, cheap wide angle lenses are used to cover a large area, but result in significant modeling errors if using the fundamental matrix approach. People typically move on 2D surfaces, such that the model assumption applies. Calibration is required to associate objects in multiple views, and should preferably be done iteratively using exclusively the incoming image sequences and without access to the scene. In practice, calibration of multi-camera surveillance systems is time-consuming and costly, as is re-calibration after disturbances such as changes of the scene or camera position (deliberate or not, for example the occasional storm can change the viewing direction several degrees). Since calibration and re-calibration add significantly to the installation and maintenance costs of such systems, there is a real need for online learning. In mobile platform applications, information from several subsystems including vision sensors need to be fused. Modalities might differ, *e.g.* IR and color images, and the configuration might change due to vibrations and other mechanical impacts. The CMap adapts to changes and allows to map between sensors of different modalities.

- *M. Felsberg, F. Larsson, and J. Wiklund are with the Computer Vision Laboratory, Dept. Electrical Engineering, Linköping University, Sweden. E-mail: michael.felsberg@liu.se*
- *N. Wadströmer is with the Division of Information Systems, FOI Swedish Defence Research Agency, Sweden*
- *J. Ahlberg is with Termisk Systemteknik AB, Sweden*

### 1.1 Approach and Problem Formulation.

The problem is to find the most likely corresponding point in the second view given a query point in the first view and a sequence of detected point sets in both views. The correspondences of the points in the sets are unknown as is the projection geometry and the scene geometry. However, it is assumed that the points are moving on (piece-wise) 2D surfaces. For each pair of point sets in the sequence, PHDs are computed using channels, a basis function approach, resulting in vectors $\mathbf{w}$ (input / first view) and $\mathbf{v}$ (output / second view). Each pair $(\mathbf{w}, \mathbf{v})$ is used to iteratively update the CMap $\mathbf{C}$, based on a suitable distance measure $D$. The PHD for a query point in the first view is then fed into the CMap $\mathbf{C}$ to estimate the PHD of the corresponding point in the other view. From the latter PHD, the most likely corresponding point is computed. Key features of this scheme are convergence and accuracy of results but also that it is computationally tractable. This requires a choice of $D$ that allows incremental updating of $\mathbf{C}$ without storing previous data, *i.e.*, online learning.

### 1.2 Related work

Most approaches for solving point correspondence problems exploit knowledge about the geometry in a statistical approach such as RANSAC [3]. Different from the proposed method, those approaches assume a perspective projection model such that the correspondence problem becomes a homography estimation problem (plane surface case) or a fundamental matrix estimation problem (general 3D case with unknown extrinsic parameters). Similar work also exists for certain, non-perspective projection models [4], but all geometric approaches have in common that the projection model needs to be known (up to its parameters) and the camera distortion needs to be estimated first.

Appearance based methods find correspondences through matching the views of objects [5]. Besides the usual problems of view-based matching, *e.g.* scale, in-plane rotation, illumination, partial occlusion etc., objects might also have different visual appearances in different images. In cases of entirely different sensors, *e.g.* an IR camera and a color camera, view matching is no alternative. Also perceptual aliasing or the existence of a multiplicity of identical objects may cause wrong correspondences.

If neither a particular projection model nor visual similarity can be assumed, learning point correspondences from unlabeled sets of feature points is an *acute* problem [6], p. 238. Since the learning data consists of sets instead of single points, this learning problem is called *correspondence-free* [7]. For unknown correspondences, all mappings are potentially multimodal (many-to-many) and a multimodal regression problem has to be solved. In [8] two online algorithms for solving this problem are proposed, one is similar to [9], based on updating covariance matrices, the other is based on stochastic gradient descent which avoids the inversion of the auto-covariance matrix, but ignores the previous error terms. Both algorithms fulfill the requirement for online learning, *i.e.* not to store the original learning data [10].

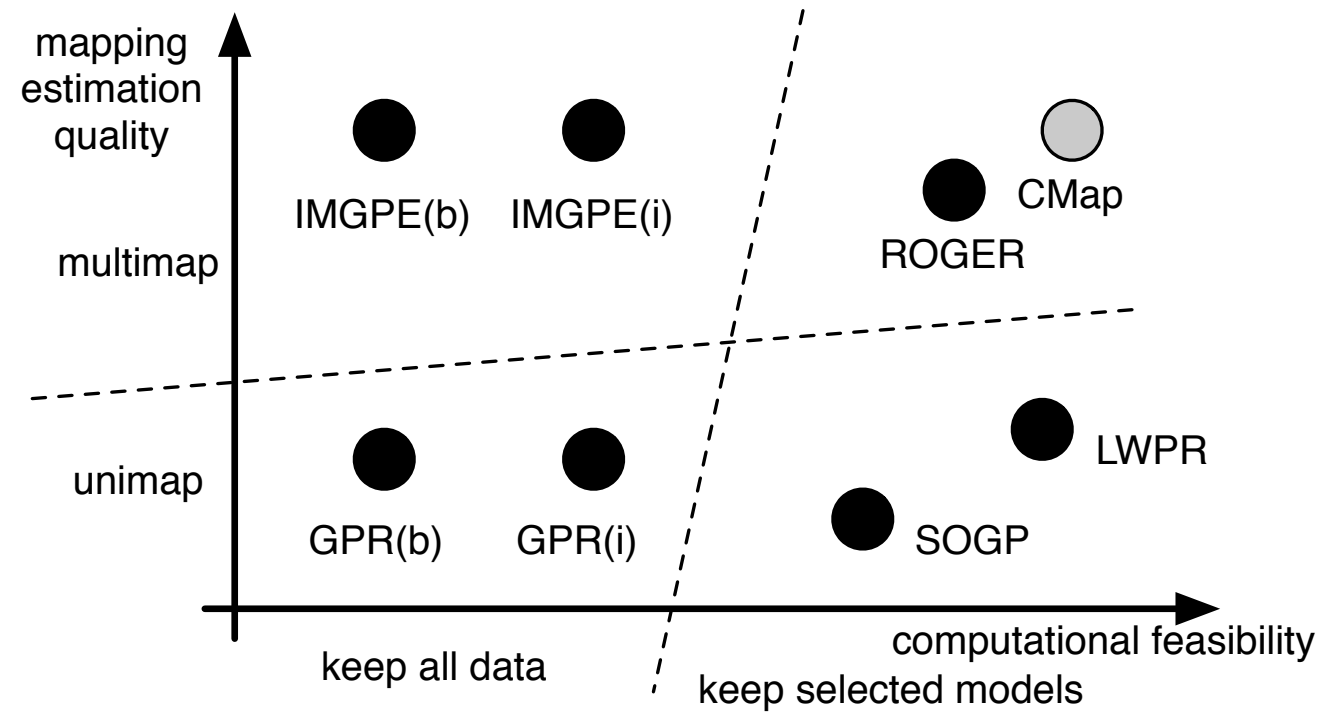

Fig. 2. Overview of learning algorithms, redrawn from [11], with the proposed method (CMap) added.

A further method for online learning of multimodal regression has been reported in the literature, called ROGER (Realtime Overlapping Gaussian Expert Regression) [11], [12]. ROGER is based on a mixture of Gaussian processes that are learned in a sparse online method [13], where sparsity means that the learning samples are reduced to a subset, *i.e.*, a property which is already covered in the previous definition of online learning (right hand-side of Figure 2). The learning is performed in two steps: assigning data points to the suitable process using a particle filter and learning the assigned process using a Bayesian approach for updating the mean and kernel functions. According to [11] ROGER is the only multimodal (multimap) online learning method, see upper right of Figure 2. If ROGER is restricted to a single Gaussian process, single samples or samples with correspondence are assumed, then ROGER becomes equivalent to SOGP (Sparse Online Gaussian Process) learning [14], placed in the lower right of Figure 2.

A popular technique for the single sample case (lower right of Figure 2) is Locally Weighted Projection Regression (LWPR), which is an incremental online method for non-linear function approximation in high dimensional spaces. The method is based on a weighted average of locally weighted linear models that each utilize an incremental version of Partial Least Squares. Each local model is adjusted through PLS to the local dimensionality of the manifold spanned by the input data. The position and size of the local models are update automatically as well as the numbers of models. LWPR has sucessfully been used for learning robot control [15], [16].

A comprehensive overview of the application field of cross-camera tracking is given in [17], where the field is divided into overlapping vs. non-overlapping views and calibrated cameras vs. learning approaches. The calibration-free approaches for non-overlapping cameras often rely on descriptors of tracked objects and proba-bilistic modeling [18], [17], where the latter requires a notion of temporal continuity. Although this continuity is not stringent for the case of overlapping cameras, known approaches make use of intra-camera tracking before establishing the correspondence [19], [20]. In the here proposed method, correspondence between uncalibrated, overlapping cameras is established on single frames and without using appearance descriptors.

### 1.3 Main contributions

The main contribution of the present paper in relation to the previously mentioned methods is a *novel algorithm for iterative learning* of a linear model $\mathbf{C}$ on a sequences of PHDs, represented as channel vectors $\mathbf{w}$ and $\mathbf{v}$, with the following properties:

- A theoretically appropriate distance measure $D$, Neyman's chi-square divergence, is proposed for learning.
- The algorithm is a proper online learning method that stores previously acquired information in the fixed size model $\mathbf{C}$ and in $\mathbf{\Omega}$, the empirical density of $\mathbf{v}$.
- The update of $\mathbf{C}$ is computed efficiently from point-wise operations on the previous data in terms of $\mathbf{C}$ and $\mathbf{\Omega}$ and the weighted residual of the incoming data.
- The algorithm has been applied to several surveillance datasets, and it shows throughout very good accuracy, produces very few outliers for the position estimates, and runs in real-time.
- On standard benchmarks, CMap learning results in very low residuals at a very low computational cost.

The paper is structured as follows: This introduction is followed by a section on the required methods, a section on the experiments and results, and a concluding discussion.

## 2 Methods

This section covers the required theoretical background of this paper: Density estimation with channels, i.e., basis functions, CMap estimation using these density estimates, the CMap online learning algorithm, and finally how to find correspondences using CMap.

In what follows, capital bold letters are used for matrices, bold letters for vectors, and italic letters for scalars. The elements of an $M \times N$ matrix $\mathbf{A}$ are given by $a_{mn}$, $m = 1 \dots M$, $n = 1 \dots N$. The elements of an $M$-dimensional vector $\mathbf{a}$ are given by $a_m$, $m = 1 \dots M$. The $n$-th column vector of $\mathbf{A}$ is given as $[\mathbf{A}]_n$, $n = 1 \dots N$. That means, the $m$-th element of $[\mathbf{A}]_n$ is $a_{mn}$. A subindex to a vector or matrix denotes a time index, *i.e.*, $\mathbf{A}_T$ is the matrix $\mathbf{A}$ at time $T$. The Hadamard (element-wise) product of matrices $\mathbf{A}$ and $\mathbf{B}$ is denoted as $\mathbf{A} \circ \mathbf{B}$ and the element-wise division is denoted as the fraction $\frac{\mathbf{A}}{\mathbf{B}}$ such that $\frac{\mathbf{A}}{\mathbf{B}} \circ \mathbf{B} = \mathbf{A}$.

### 2.1 Representations of PHDs with channels

Channel representations have originally been suggested without explicit reference to density estimation [2], [21]. In neighboring fields, these types of representations are known as population codes [22], [23]. This section introduces channel representations as described in [2], [24] for the purpose of density estimation with a signal processing approach [25].

The channel-based method for representing the PHDs is a special type of 2D soft-histogram, *i.e.*, a histogram where samples are not exclusively pooled to the closest bin center, but to several bins with a weight depending on the distance to the respective bin center. The function to compute the weights, the *basis function* $b(\mathbf{x})$, is usually non-negative (no negative contributions to densities) and smooth (to achieve stability). For computational feasibility, basis functions have compact support. In order to obtain a position independent contribution of samples to the PHD, the sum of overlapping basis functions must be constant.

The basis functions are located on a 2D grid with spacings $d_1$ and $d_2$ in $x_1$ and $x_2$. Throughout this paper the basis functions have a support of size $3d_1 \times 3d_2$. Figure 3 illustrates the cases of two and three feature detections represented in this way. Note that the basis functions adjacent to the image area also respond in some cases. The regular placement of basis functions has the major advantage that signal processing methods can be used to manipulate these representations.

In order to reconstruct the exact position of single samples or to extract modes from the channel representation, the basis-functions need to be of a particular type. Restricting soft-histograms to reconstructable frameworks shares similarities with wavelets restricting bandpass filters to perfect reconstruction. Channel representations are similar to Parzen window or kernel density estimators, with the difference that the latter are not regularly spaced but placed at the incoming samples.

For the remainder of this paper, $\cos^2$ basis functions are chosen, mainly because they have compact support (in contrast to Gaussian functions):

$$b(\mathbf{x}) \triangleq \prod_{j=1}^{2} \frac{2}{3d_j} \begin{cases} \cos^2(\frac{\pi x_j}{3d_j}) & |x_j| < \frac{3d_j}{2} \\ 0 & |x_j| \geq \frac{3d_j}{2} \end{cases} , \qquad (1)$$

where $\mathbf{x} = (x_1, x_2)^\mathsf{T}$ denotes the image coordinates. The range of $\mathbf{x}$ together with $d_j$ determine the number of basis functions $N_j = (\max(x_j) - \min(x_j))/d_j + 2$ for each dimension. The spacing $d_j$ is typically much larger than 1, resulting in $N_j$ being much smaller than the range of $x_j$. $N = N_1 N_2$ is the total number of basis functions. The 2D grid index $(k_1, k_2) \in \{0 \dots N_1 - 1\} \times \{0 \dots N_2 - 1\}$ is converted into a linear index $n(k_1, k_2)$ in the usual way: $n(k_1, k_2) = k_1 + N_1 k_2$. Using (1), the components of the $N$-dimensional *channel vector* $\mathbf{w}$ are obtained from $I$

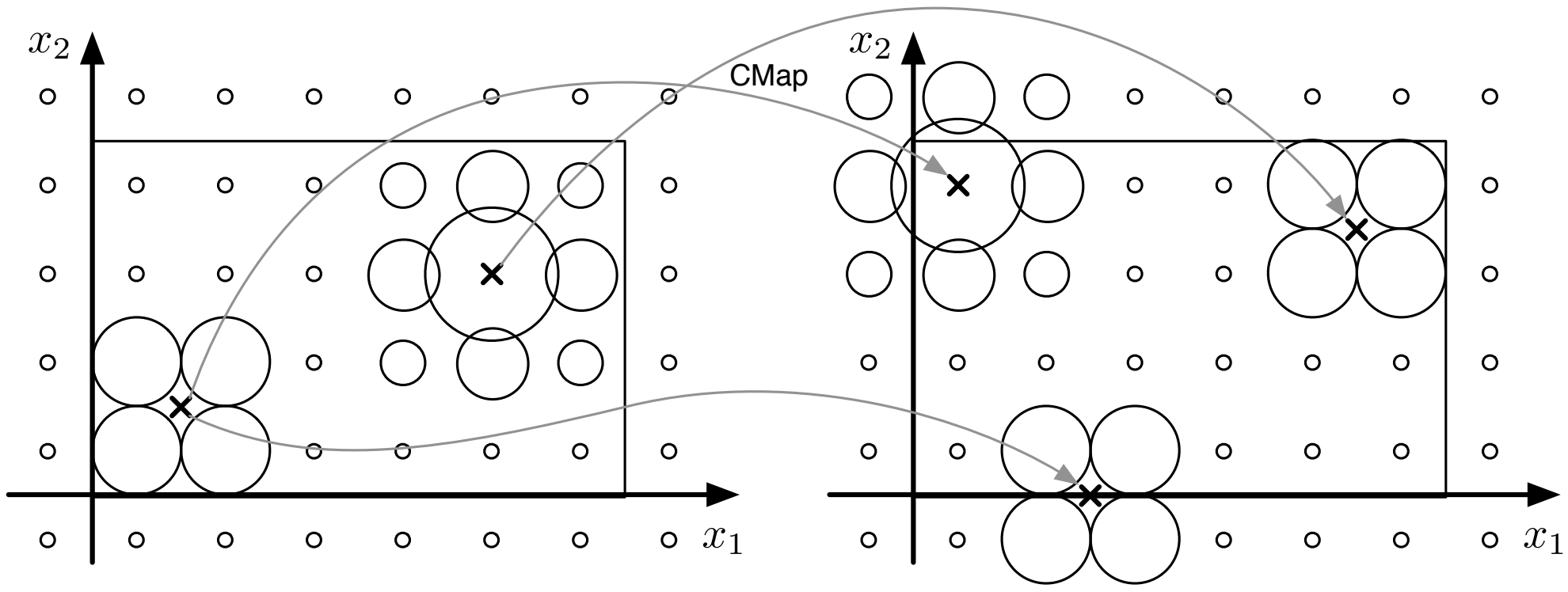

Fig. 3. The left image contains 2 detections represented using basis functions. The right one contains 3 detections. The CMap generates the right representation from the left one. The size of the circles indicates the response of the respective basis function.

image points $\mathbf{x}_i$ as

$$w_{n(k_1,k_2)} = \sum_{i=1}^{I} b(\mathbf{x}_i - (k_1 d_1, k_2 d_2)^\mathsf{T} - (d_1, d_2)^\mathsf{T}/2) \qquad k_1 = 0 \dots N_1 - 1\,,\; k_2 = 0 \dots N_2 - 1 \;. \quad (2)$$

Switching to smooth bins reduces the quantization effect compared to ordinary histograms by a factor of up to 20 and thus allows us to either reduce computational load by using fewer bins, or to increase the accuracy for the same number of bins, or a mixture of both.

In order to compute the most likely position represented by a channel vector, an algorithm for extracting the mode with maximum likelihood is required. For $\cos^2$-channels an optimal algorithm in least-squares sense is obtained as [26]

$$\hat{x}_1 = d_1 l_1 + \frac{3d_1}{2\pi} \arg\Big(\sum_{k_1=l_1}^{l_1+2} \sum_{k_2=l_2}^{l_2+2} w_{n(k_1,k_2)} e^{\frac{i2\pi(k_1-l_1)}{3}}\Big) \quad (3)$$

$$\hat{x}_2 = d_2 l_2 + \frac{3d_2}{2\pi} \arg\Big(\sum_{k_1=l_1}^{l_1+2} \sum_{k_2=l_2}^{l_2+2} w_{n(k_1,k_2)} e^{\frac{i2\pi(k_2-l_2)}{3}}\Big) \quad (4)$$

where $\arg()$ is the complex argument and the indices $l_j$ indicate the maximum sum $3 \times 3$ block:

$$(l_1, l_2) = \operatorname*{argmax}_{(m_1,m_2)} \sum_{k_1=m_1}^{m_1+2} \sum_{k_2=m_2}^{m_2+2} w_{n(k_1,k_2)} \;. \quad (5)$$

In algorithmic terms, firstly the channel vector is reshaped to a 2D array, which is then filtered with a $3 \times 3$ box filter. The maximum response defines the decoding position $(l_1, l_2)$. Secondly, within the $3 \times 3$ window at this position, the 2D position of the maximum is computed from the channel coefficients using (3–4).

### 2.2 CMap estimation using PHDs

Given two images, the detected feature points $\mathbf{x}_i$ and $\mathbf{y}_j$ are represented as channel vectors $\mathbf{w}$ and $\mathbf{v}$ respectively. For a sequence of $T$ image pairs, these vectors form matrices $\mathbf{W} = [\mathbf{w}_1 \dots \mathbf{w}_T]$ and $\mathbf{V} = [\mathbf{v}_1 \dots \mathbf{v}_T]$ and the goal is now to estimate the CMap from these matrices. The CMap is multimodal and non-continuous in general. For multimodal or non-continuous mappings, ordinary function approximation techniques, such as (local) linear regression [27], tend to average across different modes or discontinuities. Hence, Johansson *et al.* propose to use mappings from $\mathbf{w}$ to $\mathbf{v}$ instead [28]. This is also the main difference to kernel regression and the Parzen-Rosenblatt estimators, where the kernel functions on the output side are integrated to compute the estimate of $\mathbf{y}$ (see [29], p. 295). An estimate of the vector $\mathbf{v}$ is obtained using a linear model $\mathbf{C}$

$$\hat{\mathbf{v}} = \mathbf{C}\mathbf{w} \quad (6)$$

and the final estimate $\hat{\mathbf{y}}$ is obtained from $\hat{\mathbf{v}}$ using (3–4). Typical sizes of $\mathbf{v}$ and $\mathbf{w}$ are about 500 elements, such that $\mathbf{C}$ is a matrix with about 250,000 elements. Efficiency is achieved by exploiting sparseness: $\mathbf{C}$ contains typically about 5,000 non-zero elements. If the single hypothesis case is considered, all entities have corresponding probabilistic interpretations, in particular $\mathbf{C}$ corresponds to a conditional density. In the general case, however, the channel vectors correspond to combined densities of several hypotheses and the direct relation to a probabilistic interpretation is lost.

Our approach to learn $\mathbf{C}$ shares some similarities with the method of Johansson *et al.*, but is based on a different objective function. In [28] $\mathbf{C}$ is computed by minimizing the Frobenius norm of the estimation error matrix

$$\mathbf{C}^* = \operatorname*{argmin}_{\mathbf{C}} \|\mathbf{V} - \mathbf{C}\mathbf{W}\|_F^2 \quad \text{s.t. } c_{mn} \geq 0 \;, \quad (7)$$

where the non-negativity constraint leads to a reduction of oscillations and to sparse results. Note that linear regression in the channel domain allows for general (non-linear), multimodal regression in the original domain.

Since $\mathbf{C}$ acts on PHDs, columns of $\mathbf{V}$ and $\mathbf{W}$ can be added in (7) if the corresponding detections do not interfere, *i.e.*, their respective distance is larger than $3d_j$ in at least one dimension $j$, and if the detected points move independently in the learning set. Learning from the combined channel vectors has been shown to be equivalent to stochastic gradient descent on the separated vectors [7]. Further details are given in Section 2.4.

Using the Frobenius norm in (7) corresponds to a least-squares approach, which is not very suitable if channel vectors are interpreted as density representations. Also in practice, the least-squares approach causes problems with oscillating values if the non-negativity constraint is not imposed. This changes if $\mathbf{v}$ and $\hat{\mathbf{v}} = \mathbf{Cw}$ are compared using $\alpha$-divergences $D_\alpha$ (see [30], Chapter 4), as often used in non-negative matrix factorization (NMF, see [31]):

$$D_\alpha[\mathbf{v}||\mathbf{Cw}] = \sum_{i=1}^{I} \frac{\alpha v_i + (1-\alpha)[\mathbf{Cw}]_i - v_i^\alpha [\mathbf{Cw}]_i^{1-\alpha}}{\alpha(1-\alpha)} \quad . \tag{8}$$

From a theoretical point of view, divergences are a sound way to compare densities. Certain choices of $\alpha$ result in common divergence measures [32], [33]. $\alpha \to 1$ and $\alpha \to 0$ result in the Kullback-Leibler divergence and the log-likelihood ratio, respectively. $\alpha = 0.5$ results in the Hellinger distance, $\alpha = 2$ in the Pearson's chi-square distance, and $\alpha = -1$ in the Neyman's chi-square distance:

$$D_{-1}[\mathbf{v}||\mathbf{Cw}] = \frac{1}{2} \sum_{i=1}^{I} \frac{(v_i - [\mathbf{Cw}]_i)^2}{v_i} \quad . \tag{9}$$

The Neyman's chi-square is the unique $\alpha$ divergence that establishes a weighted least-squares problem. This is easily seen by requiring the numerator of (8) to be a quadratic polynomial, *i.e.*, $1 - \alpha = 2$. Furthermore, if an independent Poisson process is assumed, which is usually done for histograms, the Neyman's chi-square is equal to the Mahalanobis distance, since $v_i$ is an estimate of the variance of bin $i$. For the case of channel vectors, the bins are weakly correlated due to the overlapping basis functions and the equality only holds approximatively [22].

Hence, the Neyman's chi-square divergence is more suitable for comparing channel vectors than using the Frobenius norm in (7). Note that it is not symmetric, the denominator only contains coefficients of $\mathbf{v}$, *i.e.* empirical values. This is a conscious choice in order to stabilize iterative learning. For $T$ independently drawn samples of $\mathbf{w}$ and $\mathbf{v}$, the divergence becomes

$$D_{-1}[\mathbf{V}||\mathbf{CW}] = \frac{1}{2} \sum_{i=1}^{I} \frac{\sum_{t=1}^{T}(v_{it} - [\mathbf{CW}]_{it})^2}{\sum_{t=1}^{T} v_{it}} \tag{10}$$

and the following minimization problem is obtained

$$\mathbf{C}^* = \operatorname*{argmin}_{\mathbf{C}} D_{-1}[\mathbf{V}||\mathbf{CW}] \quad . \tag{11}$$

### 2.3 The new online CMap learning algorithm

The optimal CMap $\mathbf{C}$ according to (11) is computed iteratively from a sequence of vectors $\mathbf{w}$ and $\mathbf{v}$ using online the CMap learning algorithm.

**Algorithm 1** CMap online learning algorithm.

**Require:** $T$ instances of $\mathbf{w}_t$ are given
**Ensure:** Optimal $\mathbf{C}_t$ according to (11) and $\hat{\mathbf{v}}_t$ from (6)
1: Set forgetting factor $\gamma$, set $\mathbf{C}_0 = 0$, set $\mathbf{\Omega}_0$ as uniform
2: **for** $t = 1$ to $T$ **do**
3: **if** $\mathbf{v}_t$ is known **then**
4: $\mathbf{G} = \mathbf{C}_{t-1} \circ \mathbf{\Omega}_{t-1} + (\mathbf{v}_t - \mathbf{C}_{t-1}\mathbf{w}_t)\mathbf{w}_t^\mathsf{T}$
5: $\mathbf{\Omega}_t = \frac{(1-\gamma^{t-1})\mathbf{\Omega}_{t-1} + \mathbf{v}_t \mathbf{1}^\mathsf{T}}{1-\gamma^t}$
6: $\mathbf{C}_t = \frac{\mathbf{G}}{\mathbf{\Omega}_t}$
7: **else**
8: $\hat{\mathbf{v}}_t = \mathbf{C}_{t-1}\mathbf{w}_t$
9: $\mathbf{\Omega}_t = \frac{(1-\gamma^{t-1})\mathbf{\Omega}_{t-1} + \hat{\mathbf{v}}_t \mathbf{1}^\mathsf{T}}{1-\gamma^t}$
10: $\mathbf{C}_t = \mathbf{C}_{t-1}$
11: **end if**
12: **end for**

This algorithm is derived from the subsequent calculations. The gradient of (10) with respect to $\mathbf{C}$ gives

$$\frac{\partial}{\partial c_{mn}} D_{-1}[\mathbf{V}||\mathbf{CW}] = -\frac{\sum_{t=1}^{T}(v_{mt} - [\mathbf{CW}]_{mt})w_{nt}}{\sum_{t=1}^{T} v_{mt}} \quad , \tag{12}$$

and results in the gradient descent iteration

$$\mathbf{C} \leftarrow \mathbf{C} + \beta \frac{\mathbf{VW}^\mathsf{T} - \mathbf{CWW}^\mathsf{T}}{\mathbf{V11}^\mathsf{T}} \triangleq \mathbf{C} + \beta\, \delta\mathbf{C} \quad , \tag{13}$$

where $\mathbf{1}$ is a one-vector of suitable size and the fraction is computed element-wise (inverse Hadamard product). Adding time instance $T+1$, $(\mathbf{w}_{T+1}, \mathbf{v}_{T+1})$, the update term in (13) becomes

$$\begin{aligned} &\delta\mathbf{C}_{T+1} = \\ &\frac{[\mathbf{V}\ \mathbf{v}_{T+1}][\mathbf{W}\ \mathbf{w}_{T+1}]^\mathsf{T} - \mathbf{C}_T[\mathbf{W}\ \mathbf{w}_{T+1}][\mathbf{W}\ \mathbf{w}_{T+1}]^\mathsf{T}}{[\mathbf{V}\ \mathbf{v}_{T+1}]\mathbf{11}^\mathsf{T}} = \\ &\delta\mathbf{C}_T \circ \frac{\mathbf{V11}^\mathsf{T}}{[\mathbf{V}\ \mathbf{v}_{T+1}]\mathbf{11}^\mathsf{T}} + \frac{\mathbf{v}_{T+1}\mathbf{w}_{T+1}^\mathsf{T} - \mathbf{C}_T\mathbf{w}_{T+1}\mathbf{w}_{T+1}^\mathsf{T}}{[\mathbf{V}\ \mathbf{v}_{T+1}]\mathbf{11}^\mathsf{T}} \quad . \end{aligned} \tag{14}$$

For reasons of convergence, the old mapping is down-weighted by a forgetting factor $0 \ll \gamma < 1$ in each iteration:

$$\mathbf{C} \leftarrow \gamma\mathbf{C} + \beta\, \delta\mathbf{C} \quad , \tag{15}$$

where normalization of $\mathbf{C}$ is not necessary if applying the decoding formula (3–4).

If the update term $\delta\mathbf{C}$ is fixed for all steps and $\mathbf{C}$ has been initialized as a zero matrix, $\mathbf{C}_T$ is obtained after $T$ iterations using a geometric series ($\gamma < 1$) as

$$\mathbf{C}_T = \beta \sum_{t=0}^{T} \gamma^t \delta\mathbf{C} = \beta \frac{1-\gamma^T}{1-\gamma} \delta\mathbf{C} \quad . \tag{16}$$

Substituting $\delta\mathbf{C}_{T+1}$ and $\delta\mathbf{C}_T$ in (14) using (16) and $\beta =$

$(1-\gamma)$ (again, the absolute scale of $\mathbf{C}$ is irrelevant) gives

$$\mathbf{C}_{T+1} = \mathbf{C}_T \circ \frac{(1-\gamma^{T+1})\mathbf{V}\mathbf{1}\mathbf{1}^\mathsf{T}}{(1-\gamma^T)[\mathbf{V}\ \mathbf{v}_{T+1}]\mathbf{1}\mathbf{1}^\mathsf{T}} + \frac{(1-\gamma^{T+1})(\mathbf{v}_{T+1}\mathbf{w}_{T+1}^\mathsf{T} - \mathbf{C}_T\mathbf{w}_{T+1}\mathbf{w}_{T+1}^\mathsf{T})}{[\mathbf{V}\ \mathbf{v}_{T+1}]\mathbf{1}\mathbf{1}^\mathsf{T}} \ . \quad (17)$$

The matrix

$$\mathbf{\Omega}_T = \frac{\mathbf{V}\mathbf{1}\mathbf{1}^\mathsf{T}}{1-\gamma^T} = \frac{[\mathbf{v}_0 \dots \mathbf{v}_T]\mathbf{1}\mathbf{1}^\mathsf{T}}{1-\gamma^T} = \frac{\sum_{t=0}^T \mathbf{v}_t}{1-\gamma^T}\mathbf{1}^\mathsf{T} \quad (18)$$

denotes the cumulative $\mathbf{v}$ vector for $T$ time-steps, weighted by the forgetting factor $\gamma$ (identical in all $N$ columns). Substituting $\mathbf{\Omega}_T$ and collecting terms, the final update equation for CMap learning is obtained

$$\mathbf{C}_{T+1} = \frac{\mathbf{C}_T \circ \mathbf{\Omega}_T + (\mathbf{v}_{T+1} - \mathbf{C}_T\mathbf{w}_{T+1})\mathbf{w}_{T+1}^\mathsf{T}}{\mathbf{\Omega}_{T+1}} \ . \quad (19)$$

As it can be seen in the CMap algorithm, $\mathbf{C}$ is only updated if $\mathbf{v}_t$ is known, *i.e.* learning data is available. Otherwise, only $\mathbf{\Omega}$ is updated and an estimate of $\mathbf{v}$ is computed. The latter is always used during evaluation.

### 2.4 Properties of CMap learning

The proposed CMap update equation differs from the two previously suggested schemes [8] for solving (7). The first scheme involves computing the auto-covariance matrix of $\mathbf{w}$ and the cross-covariance matrix of $\mathbf{w}$ and $\mathbf{v}$. It is suggested to incrementally update these covariance matrices by introducing a forgetting factor $\gamma$ and adding the outer product of new channel vectors. The update of $\mathbf{C}$ in time-step $T+1$ is determined by

$$\gamma(\mathbf{C}\mathbf{W}_T - \mathbf{V}_T)\mathbf{W}_T^\mathsf{T} + (\mathbf{C}\mathbf{w}_{T+1} - \mathbf{v}_{T+1})\mathbf{w}_{T+1}^\mathsf{T} \quad (20)$$

up to a point-wise weighting factor.

In the second scheme, the first term in (20) is assumed to tend to zero for large $T$, thus (20) results in a stochastic gradient descent based on the second term only. Thus, both schemes differ from (19) in the way how old and new information is fused: Scheme 1 keeps updating using the old covariance matrices whereas scheme 2 ignores old data altogether. The proposed CMap scheme re-weights the previous matrix $\mathbf{C}$ point-wise with $\mathbf{\Omega}_T$ and weights the combined update with $\mathbf{\Omega}_T^{-1}$, which results in an exact minimizer of Neyman's chi-square distance.

In contrast to the approach in [28], the CMap $\mathbf{C}$ may contain negative values as long as the product $\mathbf{C}\mathbf{w}$ is non-negative for all valid channel vectors $\mathbf{w}$. Valid channel vectors are formed by applying the basis functions (1) to a (continuous) distribution. The basis functions have a low-pass character, which implies that the subsequent operator $\mathbf{C}$ may be of high-pass character, as long as the joint operator is non-negative. High-pass operators typically have small negative values in the vicinity of large positive values. In practice, the update (19) sometimes results in negative values and to avoid problems with stability, large negative values are clipped to small negative values.

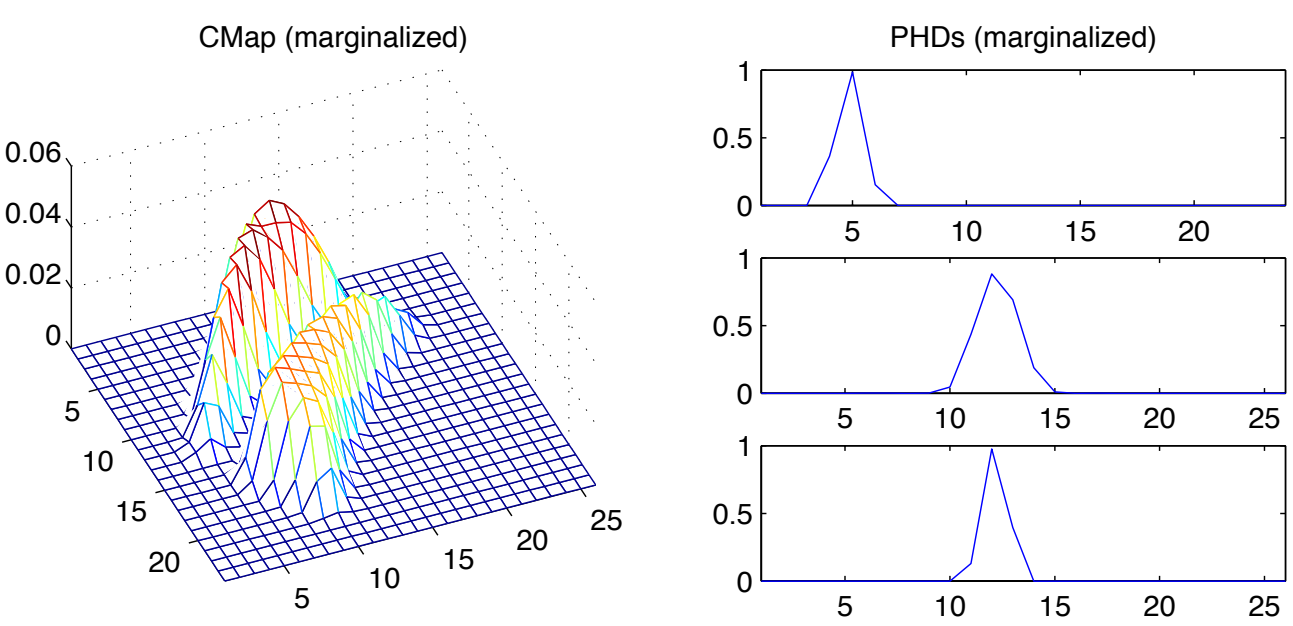

Fig. 4. Illustration of multimodal CMap (left) learned from step data (Figure 6, middle) and example PHDs (right); from top to bottom: Input PHD, resulting output from CMap, ground truth PHD. All densities are marginalized by integrating out the $y$-coordinate.

In the terminology of online learning [10], the update equation (19) is a proper online learning method, as no learning data is accumulated through time. All involved entities are of strictly constant size, and therefore, there are basically no limitations in the duration of the learning. In practice, (19) has been applied for hours in real-time applications within the DIPLECS[1] project. In this project, the same learning method has also been used on other types of problems than camera-to-camera mapping, in particular on learning tracking in non-perspective cameras [34].

Another fundamental property of CMap learning is its capability to find the respectively simplest mapping. If the problem to be learned is unimodal, the learned CMap will result in unimodal output, see Figure 4. The reason for this behavior is the equivalence to learning with correspondences [7], also observed for the theoretical case of general PHDs [1]: *"... time evolution of the PHD is governed by the same law of motion as that which governs the between-measurements time-evolution of the posterior density of any single target ..."*.

### 2.5 Finding correspondences using CMap

Once the CMap $\mathbf{C}$ has been learned, it can be used in two different ways: In *position mode* and in *correspondence mode*.

In position mode, no output feature points $\mathbf{y}_j$ are known. Each input feature point $\mathbf{x}_i$ is separately encoded as a channel vector $\mathbf{w}_i$ and mapped to the output channel vector using (6)

$$\hat{\mathbf{v}}_i = \mathbf{C}\mathbf{w}_i \ . \quad (21)$$

The output channel vector is then decoded using (3–4) to obtain output feature coordinates $\hat{\mathbf{y}}_i$. The confidence of this decoded point is given in terms of the sum of coefficients in (5). If this confidence is below

1. http://www.diplecs.eu/

a certain threshold, the input point $\mathbf{x}_i$ is considered to being mapped outside the considered range or to having no corresponding point. If the confidence of the first decoding (mode) is above the threshold, further modes of $\hat{\mathbf{v}}_i$ could be decoded, but this has not been considered in the subsequent experiments.

In correspondence mode, also the output feature points $\mathbf{y}_j$ are known, but not their correspondence. Again, each input feature point $\mathbf{x}_i$ is separately encoded as a channel vector $\mathbf{w}_i$ and mapped to the output channel vector $\hat{\mathbf{v}}_i$. Also each output feature point $\mathbf{y}_j$ is separately encoded as a channel vector $\mathbf{v}_j$. For each input feature $i$, the corresponding output feature $j$ is now found as

$$j(i) = \underset{j}{\operatorname{argmax}}\, \mathbf{v}_j^\mathsf{T} \mathbf{C} \mathbf{w}_i \ . \tag{22}$$

If the scalar product $\mathbf{v}_{j(i)}^\mathsf{T} \mathbf{C} \mathbf{w}_i$ is below a certain threshold, no correspondence to point $i$ is found. Also, some $j$ might not be assigned to any $i$, thus no correspondence to point $j$ is found. Basically, the assignment can also be computed the other way around as $i(j)$, or both ways. Combining the two-way assignment by a logical *and* results in a one-to-one assignment, combining them by a logical *or* results in a many-to-many assignment. For the experiments below, only the first case $j(i)$ has been considered.

## 3 EXPERIMENTS

CMap learning is evaluated on five different datasets D1–D5:

- D1: The cross 2D data as used in [35]
- D2: Synthetic projections of 3D points using two real camera models
- D3: Surveillance data from PETS2001
- D4: Surveillance data from the PROMETHEUS dataset
- D5: Surveillance data acquired at a spiral staircase

All these datasets have been selected according to the original problem formulation in Section 1.1: The points are located on 2D surfaces.

The CMap algorithm is compared to LWPR [35] and ROGER [12] on the dataset D1. ROGER and four different update schemes for CMap are compared quantitatively using the dataset D2. A quantitative comparison to ROGER and a convergence analysis are shown on the dataset D3. Finally, qualitative evaluation of the CMap learning is made on D4 and D5, where in particular the latter shows that CMap learning is not restricted to the planar case, but also works on real data for curved surfaces. The parameters for the comparisons between ROGER and CMap are summarized in Table 1.

| method | parameter | D1 | D2 | D3 |
|---|---|---|---|---|
| ROGER | $\mu$ | 1 | 0.5 | 0.5 |
| | $\lambda$ | 2 | 0.1 | 2 |
| | $\alpha$ | 0.5 | 0.1 | 0.5 |
| | $k$ | 0.01 | 0.01 | 0.01 |
| | $v$ | 2 | 3 | 3 |
| | $P$ | 1 | 20 | 30 |
| CMAP | $\gamma$ | $1-10^{-4}$ | $1-10^{-4}$ | $1-10^{-3}$ |
| | $N_j$ input | $33\times 33$ | cf. Table 2 | $50\times 38$ |
| | $N_j$ output | 8 | cf. Table 2 | $50\times 38$ |

TABLE 1
Overview choice of parameters for ROGER and CMap.

### 3.1 Cross 2D data (D1)

The cross 2D data is generated from a process with constant variance $\sigma^2 = 0.01$ and mean in the range $[0; 1.25]$, given by the combination of three 2D Gaussian functions [35]. One example for a set with 500 samples is shown in Figure 5, top. Sets with varying number of samples are used to train the respective method. The evaluation is done using the ground truth density on a grid of $41\times 41$ points, using the normalized mean square error (nMSE), *i.e.*, MSE divided by the variance of the noisy input data.

The code for the LWPR method and for generating the dataset are taken from Vijayakumar's homepage[2]. All parameters have been kept unchanged: Gaussian kernel with $\mathbf{D} = \begin{pmatrix} 100 & 0.01 \\ 0.01 & 25 \end{pmatrix}$ (the performance boost described in [13] by increasing $\mathbf{D}$ could not be observed), $\alpha = 250$, $w = 0.2$, meta learning set to one and the meta rate to 250.

The ROGER implementation has been taken from Grollman's homepage[3]. All parameters have been left unaltered, except for $\lambda = 2$, which gives significantly better results than $\lambda = 0.1$, cf. Table 1.

The parameters for CMap learning are also listed in Table 1. The range of the input space is $[-1; 1] \times [-1; 1]$. The range of the output space is fixed to $[-0.2; 1.35]$, which includes most of the samples from the cross 2D data.

All three methods (CMap, ROGER, LWPR), with parameters as detailed above, are evaluated on the cross 2D data with 500, 1,000, 2,000, 4,000, 6,000, 8,000, and 10,000 samples.

### 3.2 Results on cross 2D data

The result of CMap learning for 500 samples (less than 20 activations per input channel) is shown in Figure 5, middle. The normalized MSE for all three methods is plotted in Figure 5, bottom. In general, CMap learning converges fastest. The accuracy is always better than LWPR with at least a factor of 3. However, the results for LWPR from [35] could not fully be reproduced. The reported numbers are similar to the here reported LWPR results for low numbers of samples but better by a factor of two for 10,000 samples (but still less accurate than CMap learning).

2. http://homepages.inf.ed.ac.uk/svijayak/software/LWPR/LWPRmatlab/
3. http://lasa.epfl.ch/~dang/code.shtml

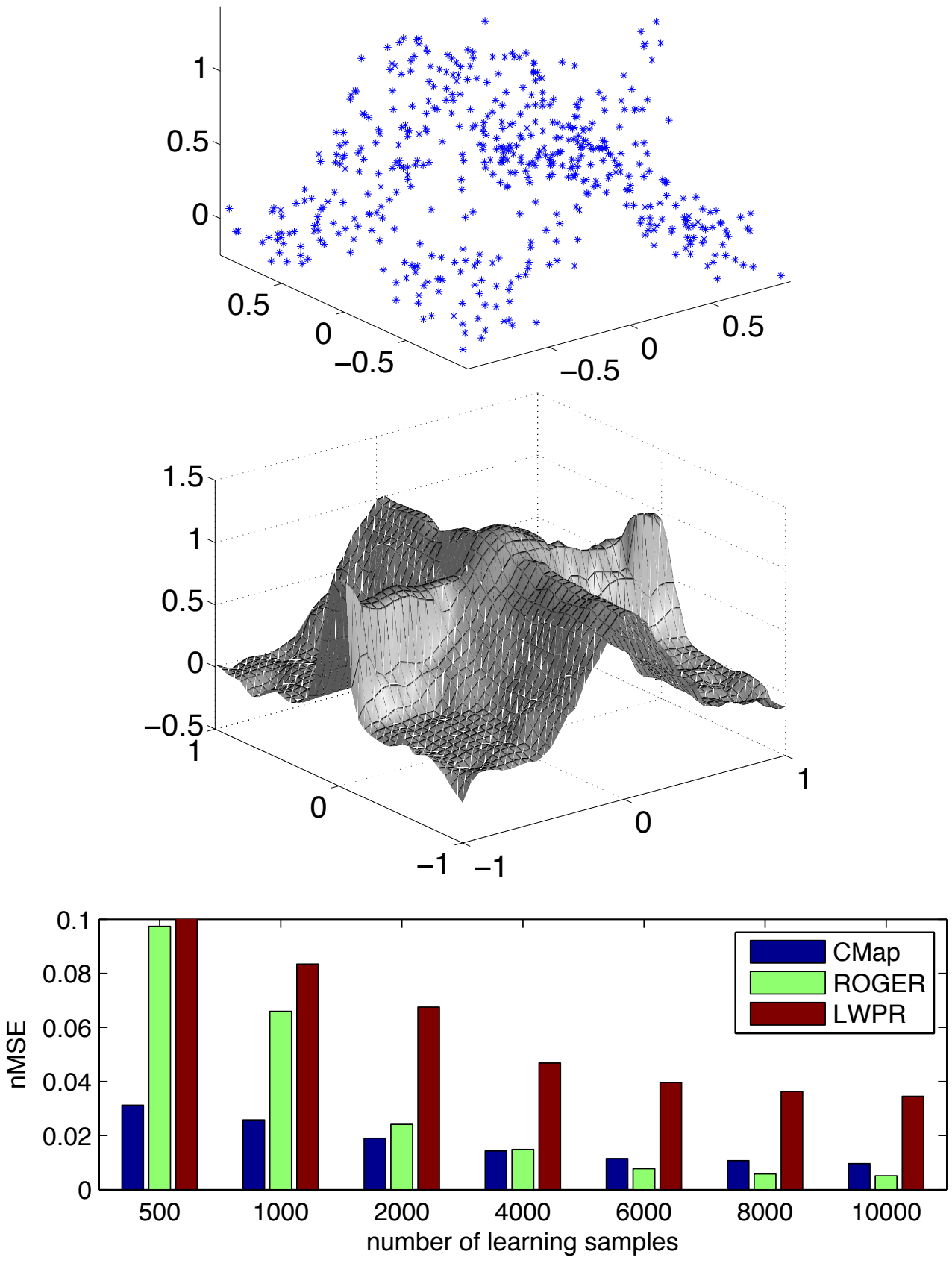

Fig. 5. Top: 500 samples from cross 2D data. Middle: Evaluation for the mapping learned using CMap learning. Bottom: Comparison with LWPR and ROGER. The barplot shows the nMSE over the number of learning samples.

Both methods run in real-time (30 fps) at about the same speed. The computation time per frame does not increase with time. This is different for ROGER: Albeit having real-time performance for the first few samples, it slows down with increasing number of incoming samples. As a consequence, the learning took about 3 hours for 4,000 samples, compared to less than a minute for the other methods.

In terms of accuracy, ROGER starts at about the same level as LWPR, but improves accuracy faster and reaches the same level of accuracy as CMap learning at 4,000 learning samples. Eventually, ROGER is about twice as accurate as CMap learning, but for the price of longer run-time by about 3 orders of magnitude. Note also that [13] reports non-normalized MSEs which cannot be compared directly to the nMSEs reported here and in the original work [35].

### 3.3 Synthetic projections dataset (D2)

This dataset consists of realistic projections of synthetic data, similar to the experiment in [6]. From a real two-camera setup, camera calibration and radial distortions are estimated and used to build a realistic projection model. This model then generates image points from a $18 \times 18$ array of 3D points. The 3D points lie on three different surfaces: A plane, a hyperbolic surface, and a two-plane discontinuity, see Figure 6.

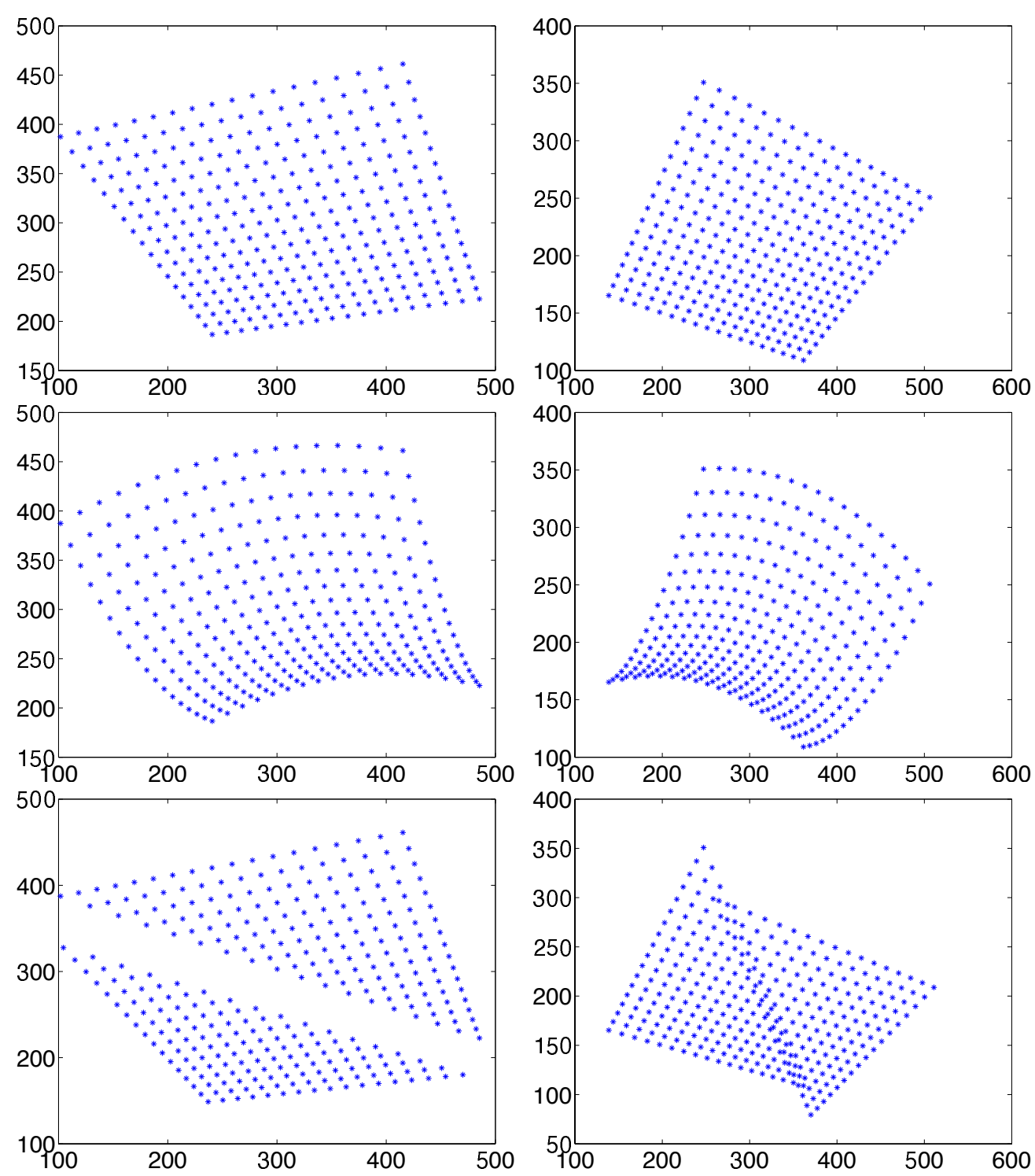

Fig. 6. Projections of synthetic 3D points. Left: Camera 1, right: Camera 2. Top: Plane surface, middle: Curved surface, bottom: Discontinuous surface.

Noise with four different standard deviations (0, 1, 2, and 3 pixel in $x_1$ and $x_2$ direction) is added to the projected points. The training samples are arranged in two ways: a) Each sample contains one point-pair, *i.e.* the point correspondence is known and only correct associations (no outliers) are in the data (datasets 'flat', 'curved', and 'step'). b) Each sample contains two points for each of both images and the correspondence is unknown. Thus four pairs are given, two of which are correct, such that 50% of the data consist of outliers (datasets 'flat2', 'curved2', and 'step2'). In total, the dataset consists of 24 different cases. The CMap is learned sequentially for each case and evaluated based on the position error of the last 9 predicted positions. The evaluation is repeated for 10 different instances for each subset to obtain statistically reliable results.

In a second evaluation, CMap learning is compared to other methods on the dataset 'flat2'. LWPR cannot deal with the multiple point data such that only ROGER is compared to four different update schemes for $\mathbf{C}$: Stochastic gradient decent [8] (called LMS), LMS with a decay factor (wLMS), the proposed update (19), and a variant of the latter without decay factor (Chi2). Since ROGER breaks down for 50% outliers, the number of outliers had to be reduced to 25% by showing every other point with known point correspondence (*i.e.*, from dataset 'flat').

| scheme | proposed | LMS | Chi2 | wLMS |
|---|---|---|---|---|
| $N_j$ input | 22 | 22 | 24 | 22 |
| $N_j$ output | 24 | 20 | 26 | 20 |

TABLE 2
Number of basis functions for the different schemes.

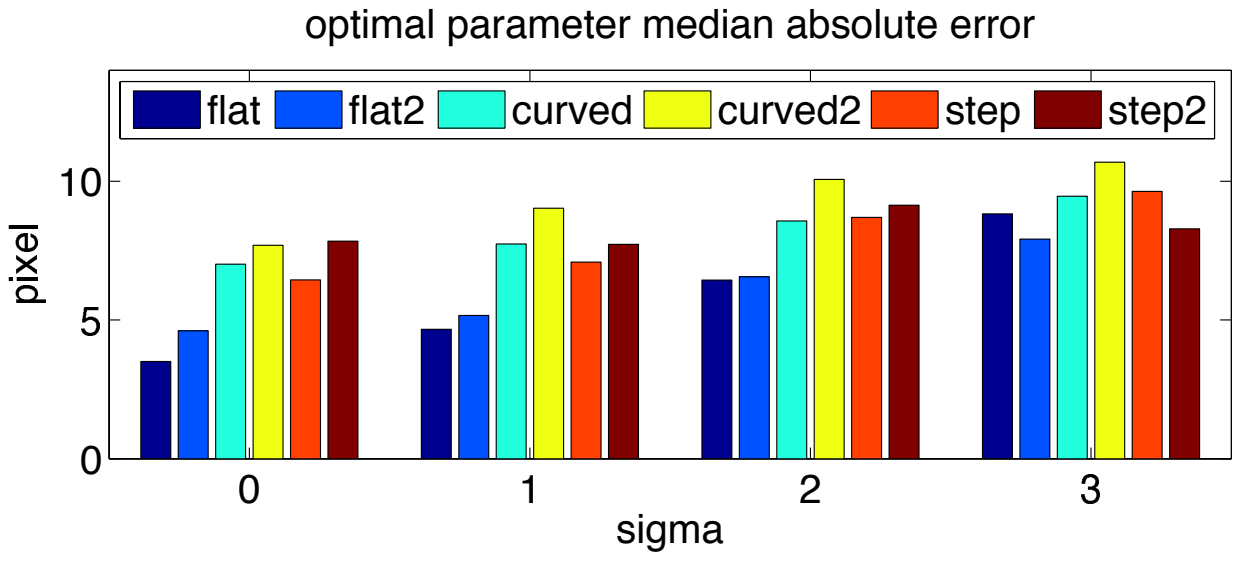

Fig. 7. D2: Localization error of the proposed scheme depending on the surface complexity, noise level, and correspondence. A '2' indicates that two points are contained in each sample.

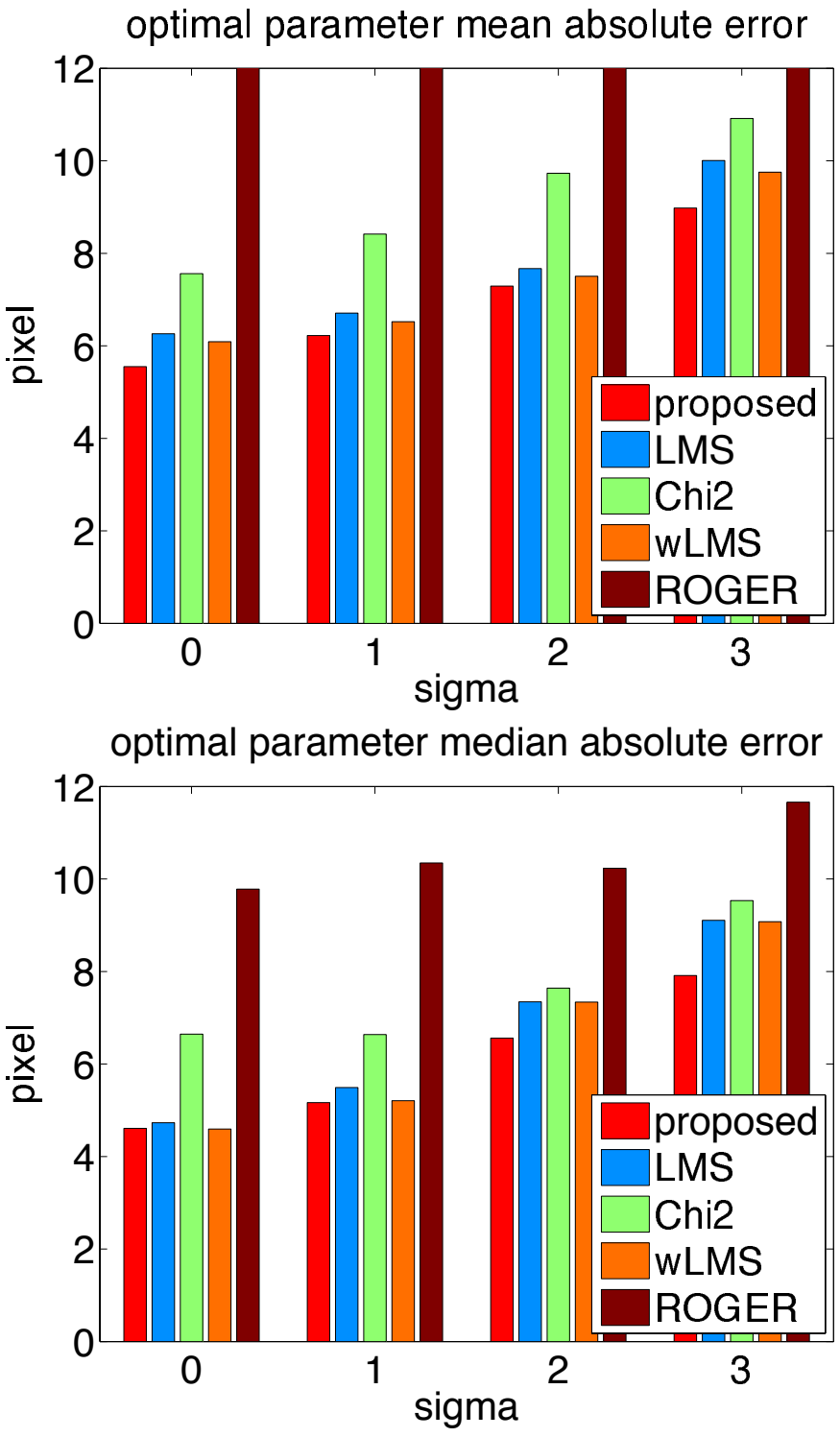

Fig. 8. Experiment on data set D2, flat2. Localization error depending on the method, plotted over the noise level. The training samples contained outliers.

The parameters for ROGER have been manually tuned to obtain low errors. The input data has been rescaled to $[-0.5; 0.5]$, the output data to $[-2.5; 2.5]$. The number of experts has been set to $P = 20$ and the other parameters have been $\alpha = 0.1$, $\mu = 0.5$, and $\lambda = 0.1$. For the proposed method, the decay factor remained as in Section 3.1. The $N_j$ have been optimized for each method over the whole dataset and for an area of $400 \times 400$ pixels, see Table 2.

### 3.4 Results on synthetic projections dataset

The median absolute position error for the CMap scheme on all 24 datasets is plotted in Figure 7. The accuracy degrades slightly for learning without correspondences and with increasing complexity of the surfaces. As expected [7], wrong correspondences are sorted out by low likelihood of consistency through the dataset. The shape of the surface becomes less relevant with increasing noise level. CMap also deals successfully with the non-unique solution at the discontinuity, due to its multimap capabilities. The overall accuracy is significantly better than the resolution of the grid (grid cells are bout $20 \times 20$ pixels), cmp. Figure 7.

For the second evaluation, the mean and median absolute error for the last 9 predictions for 'flat2' (averaged over 10 instances) are plotted in Figure 8. Consistent median and mean error indicate a low outlier rate. In most cases, the outlier rates are very low, except for ROGER. As expected the error grows with the noise level and LMS and wLMS perform equally well. The weighting only matters here if the data generating process is not stationary. The proposed scheme works best throughout, despite that its unweighted variant (Chi2) gave the poorest accuracy among the CMap optimization schemes. Although the accuracy of ROGER is slightly better than CMap for corresponding points (cf. Figure 5), its performance degrades under the presence of outliers.

### 3.5 Surveillance datasets (D3, D4, and D5)

One of the major application areas for learning CMap is multi-camera surveillance. The PETS2001 datasets[4] are a popular testbed for evaluating surveillance algorithms. Since ground truth information is needed for evaluation, TESTING dataset 1 (2688 frames) has been used here. The center positions of all clearly visible pedestrians (8 instances) have been used for comparing ROGER and CMap learning. More than 64% of the training samples contain more than one point and no information about point correspondences is given. In some cases, small groups of people move through the scene, represented by a cluster of points that behaves like a non-rigid object.

All 2688 frames are randomly re-ordered since otherwise the temporal consistency of point positions would not allow us to perform a proper cross-evaluation with the respectively subsequent samples. The point positions in cameras 1 and 2 are fed frame by frame into the respective method. For each new frame, first the positions in camera 2 are estimated for all points in camera 1. Then

4. ftp://ftp.cs.rdg.ac.uk/pub/PETS2001/

the two sets of points are added to the learning. The whole procedure has been repeated 10 times, with new re-orderings.

For ROGER, the standard parameters are used, except for $P = 30$ and the data is downscaled to the interval $[-1; 1]$. For CMap learning, the decay factor is $\gamma = 1-5\cdot 10^{-3}$ and the basis functions are placed with a spacing of 16 pixels.

CMap learning has also been applied to dataset D4, provided by the PROMETHEUS project[5], and dataset D5. Dataset D4 contains positions of persons moving in the scene that are detected with a combination of background modeling and head detection. Due to shadows it is difficult to detect the point on the ground where the persons stand. A detector of radial symmetry is used to find the heads. Dataset D5 is a novel dataset acquired in a spiral staircase from two uncalibrated views, one from the side and one from the top. Five different sequences show two people walking up and down the stairs. From these sequences, the image positions of the head centers are extracted and used for learning.

The only parameters that have been changed for D4/D5 compared to D3 are the number of channels: $50\times 32$ for D4 and $34\times 26$ for D5. The evaluation is purely qualitatively, as only the most likely correspondences between the views are indicated.

### 3.6 Results on surveillance datasets

An example for the position accuracy of CMap on PETS2001 is given in Figure 1. The quantitative results for ROGER and CMap on PETS2001 with unknown correspondence are plotted in Figure 9. For comparison, also the accuracy of normalized DLT homography estimation [3] on the respective point-sets *with* known correspondence (*i.e.* no outliers) is included. Each box illustrates the distribution of the distance to the true center; the box contains the 2nd and 3rd quantile and the red line is the median. The position estimates are pooled depending on the number of previously seen samples, *i.e.* the second box contains all estimates of the respective method after between 69 and 88 learning samples. CMap has initially a larger error but converges towards about half the error of ROGER and achieves basically the same accuracy as normalized DLT homography estimation, *despite* that the latter uses correspondence information. The convergence behavior is exactly opposite to that one observed in Figure 5, which is presumably caused by the fact that ROGER generates models covering the whole image when the first learning samples are observed. These global models seem to be maintained throughout and only if the error of the global models becomes too high, newer local models are used for the output – as it happens in the case of the cross 2D data. For the PETS2001 data, the error of the global model is apparently too low to trigger the use of new local models. CMap, on the other hand, uses solely local models and performs sub-optimal on data that perfectly suits a combination of global models (such as cross 2D data). For data with varying local behavior, CMap learning results in models with better fidelity than ROGER, but requires more learning samples to converge. According to the plots in Figure 9, ROGER has converged at about 350 samples, whereas CMap needed about twice as many.

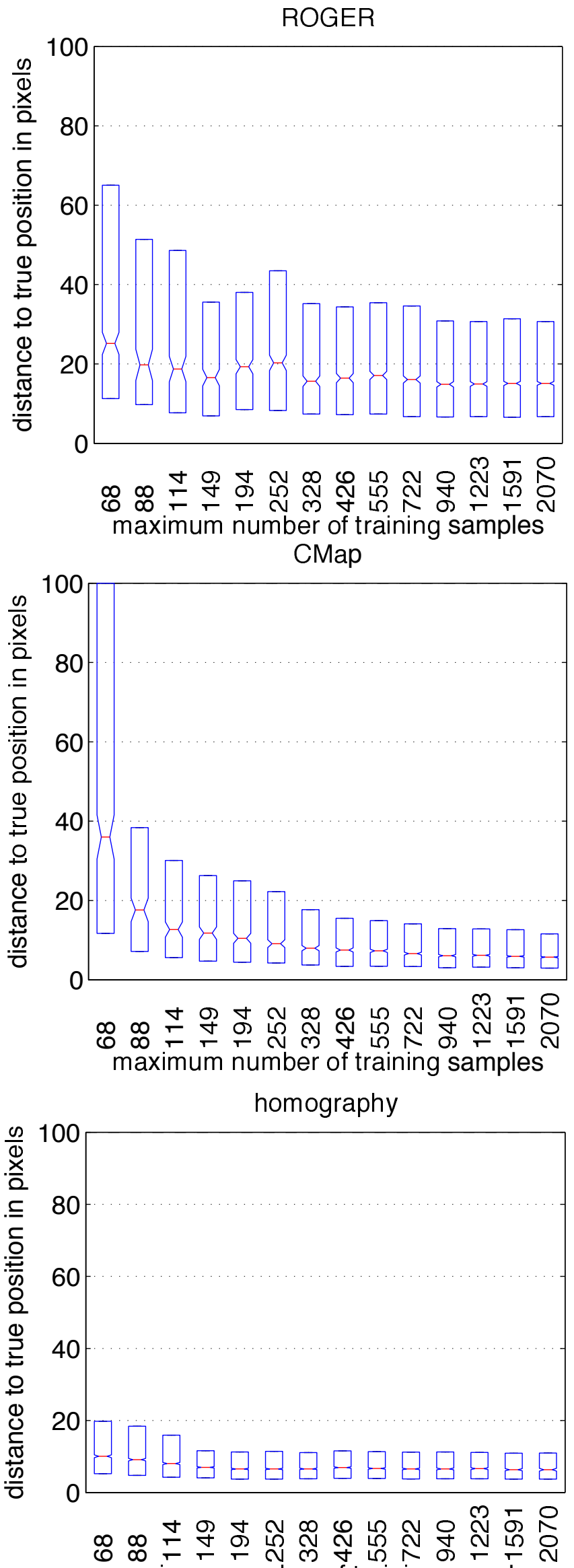

Fig. 9. Results for ROGER (top), CMap (middle), and standard homography estimation (bottom) on PETS2001. ROGER and CMap are trained without known correspondences, whereas homographies have been estimated using corresponding points.

The convergence speed of CMap depending on $\gamma$ and on the ordering of the data (ordered vs. random permutations) has been analyzed for 100 random subsets each, see Figure 10. The parameter $\gamma$ has been varied between $1-4.3\cdot 10^{-4}$ and $0.5$. Despite this wide range, results vary

5. http://www.prometheus-fp7.eu/

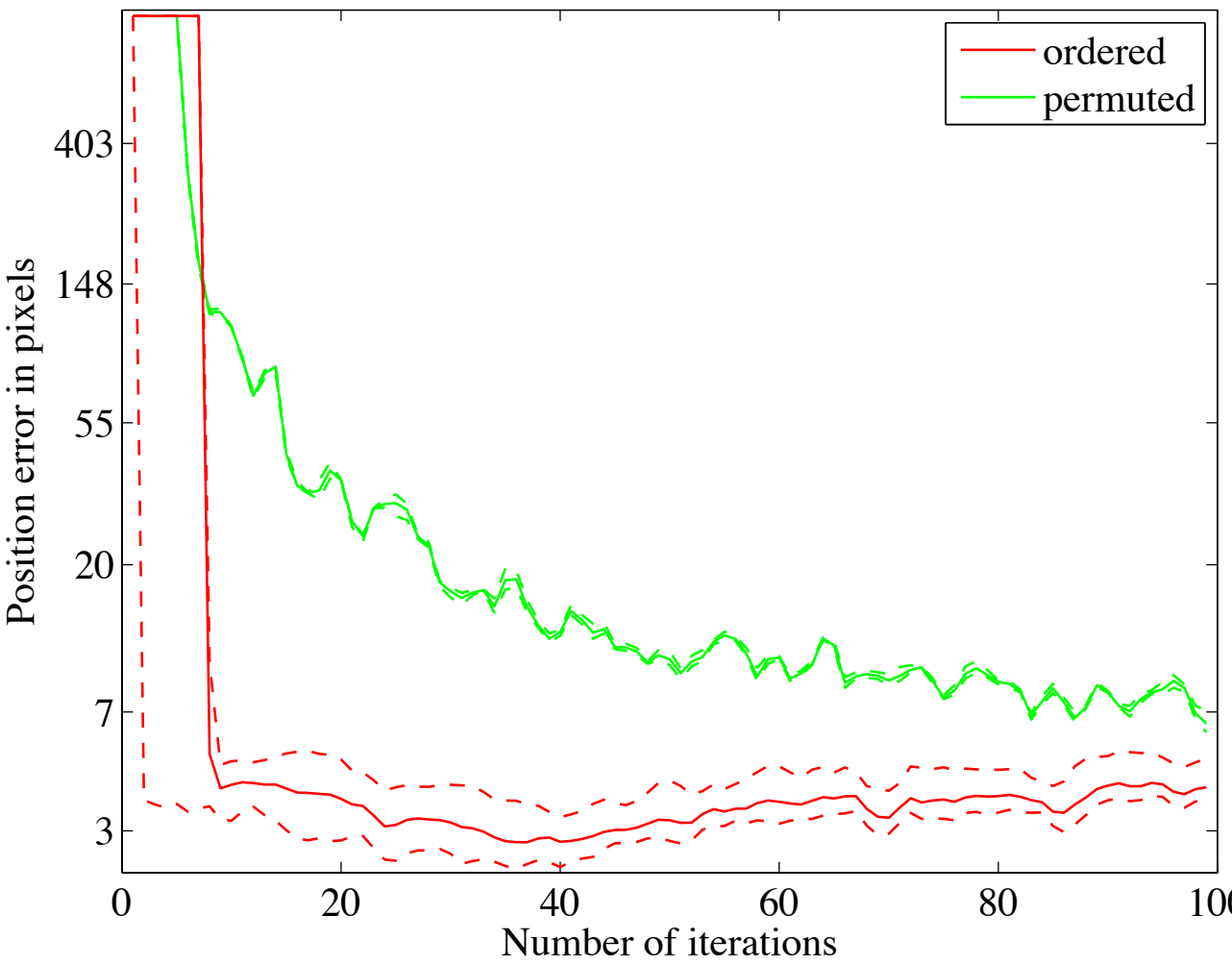

Fig. 10. Convergence for ordered data (red) and permuted data (green). The solid lines indicate the mean logarithmic error over all $\gamma$, the dashed lines the maximum and minimum logarithmic error, respectively.

only mildly and only in the ordered case, with larger values for $\gamma$ giving slightly more accurate results. The figure also shows that keeping data ordered results in highly accurate results after just a few iterations. This is caused by the locality of subsequent samples, *i.e.*, new samples are close to previously seen positions. However, this also means that the mapping initially (after about 40 iterations) only covers a small portion of the view and has a tendency to over-fit. Adding more learning samples makes the error increase slightly, but a larger area of the view is covered. Random permutations of learning samples converges slower, but covers the whole view globally right from the beginning.

On the PROMETHEUS dataset, CMap finds correct correspondences already after about 150 learning frames, see Figure 11 (a) and accompanying video. The input data contains outliers (false positive detections and false negative detections) in the input space and in the output space, but CMap learning is not affected for the same consistency argument as in Section 3.4. The only exceptions are in cases where too few examples have been observed, Figure 11 (c), or two detection outliers in close vicinity occur, Figure 11 (e). In both cases, only a single frame has been affected, which can easily be corrected by temporal filtering.

On dataset D5, CMap has been trained on four of the sequences (about 150 frames) and is evaluated on the respectively 5th sequence. CMap finds correct correspondences in nearly all frames, except for a few single frames where a correspondence is missing due to the confidence of the mapping being too low. Two frames from the data set are shown in Figure 12. The spiral staircase experiment shows that CMap learning is not restricted to planar surfaces, but also works for curved surfaces with discontinuities.

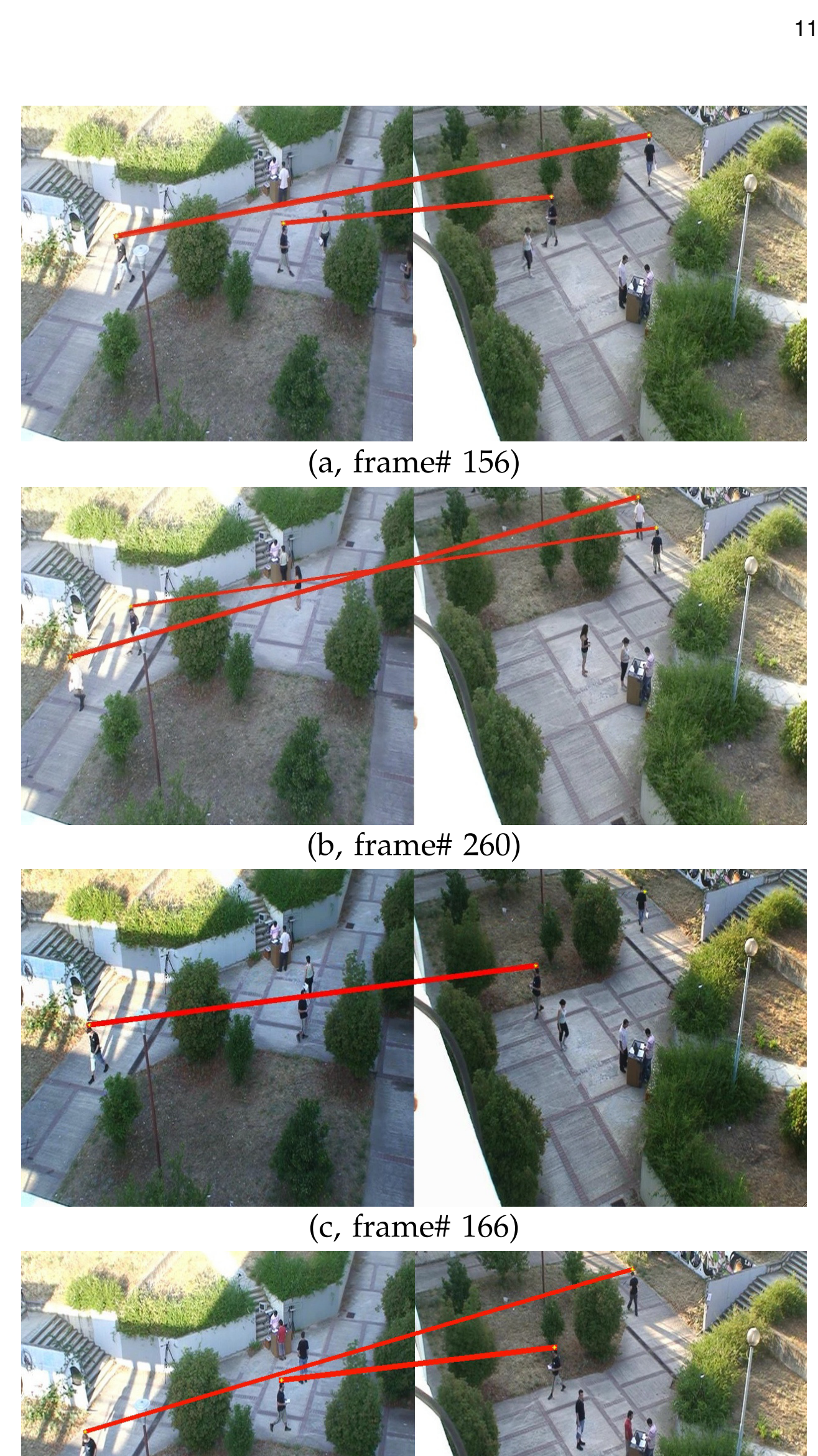

Fig. 11. Examples from the PROMETHEUS dataset D4. Correct association of detected walking persons (a, b); wrong association due to missing detection in the left view and short learning period (c); correct association in a similar situation, but longer learning period (d); wrong association due to confused detections of two close-by persons (e). For further examples, see video.

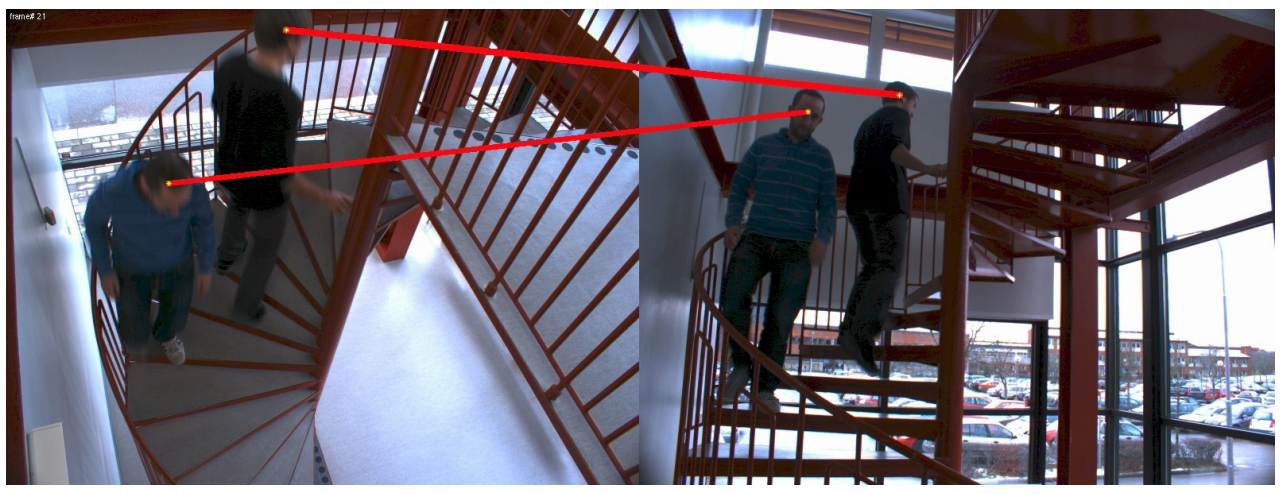

(a, frame# 21, sequence# 1)

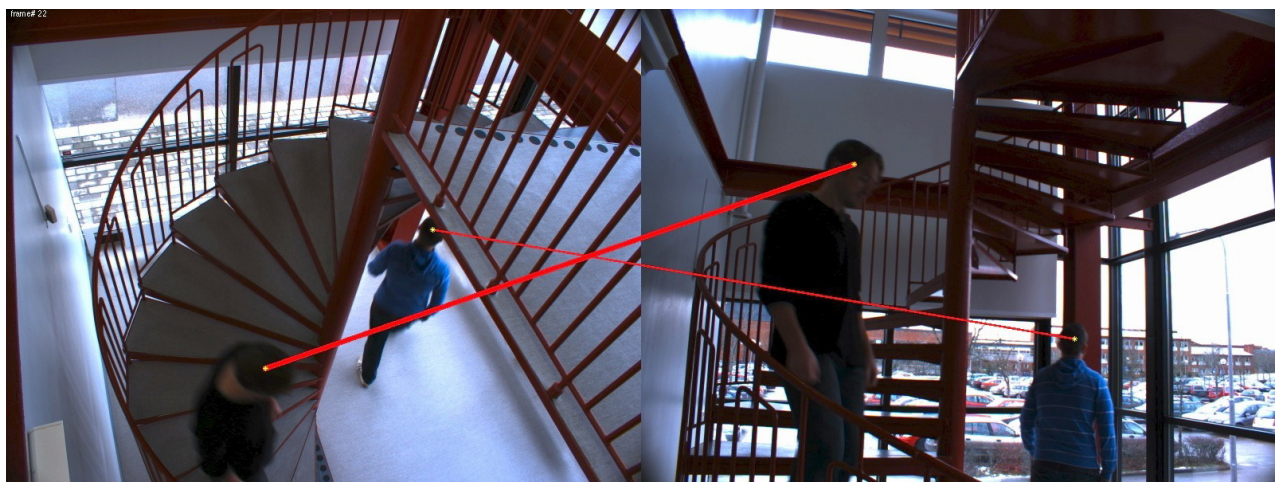

(b, frame# 22, sequence# 4)

Fig. 12. Examples from the spiral staircase dataset D5. For further examples, see video.

## 4 Conclusion

A new method for online learning of the CMap between images has been introduced in this paper. Neyman's chi-square divergence is proposed for learning the linear model $\mathbf{C}$ on density representations. This distance measure is theoretically appropriate as it is designed to compare densities, *i.e.*, non-negative functions, in contrast to the least-squares distance. A new iterative algorithm is derived, which is a proper online learning method and stores all previous data in the fixed size model $\mathbf{C}$ and a fixed size vector $\mathbf{\Omega}$. Compared to stochastic gradient descent, information from previous data is kept in terms of $\mathbf{\Omega}$. The update of $\mathbf{C}$ is computed efficiently from point-wise operations on the previous data in terms of $\mathbf{C}$ and $\mathbf{\Omega}$ and the weighted residual of the incoming data. Its efficiency makes the algorithm very suitable for real-time systems. CMap learning has been applied to several surveillance datasets, and it shows better accuracy and produces fewer outliers for position estimates than other updating schemes for $\mathbf{C}$, *e.g.* LMS. On standard benchmarks, the proposed method is significantly more accurate than LWPR and ROGER at the same computational cost. Only with parameter setting leading to prohibitively high computation efforts, ROGER achieves marginally better accuracy. On the PETS2001 dataset, CMap learning shows faster convergence and smaller variance of results than ROGER and is more robust against outliers.

The main limitation of CMap learning is the resolution of the underlying channel representation. If two objects are closer than what can be distinguished by the channel representation, they might be confused. The position accuracy is also limited by the width of the basis functions, although it is an order of magnitude better than the spacing of the channels. On the other hand, the resolution of the channel representations should not be chosen too high, since learning will then degenerate to reproduce the noise in the learning data and the computational burden increases unnecessary. As a rule of thumb, placing channels in a regular grid with about 20 pixels spacing is a good compromise between accuracy, confusion, noise suppression, and generalization capabilities. Further problems might occur if different objects move highly correlated. In that case, CMap learning potentially confuses these objects.

The algorithm used for CMap learning is not restricted to 2D correspondence problems. However, correspondence learning is a good illustration of the properties and advantages of the proposed method. The same method has also been used for replacing the learning algorithm for tracking problems [34] within the DIPLECS[1] project. A video demonstrating the tracking results, among other DIPLECS achievements, is available at the project website. Future work will concentrate on applying the learning algorithm on combined mapping and tracking problems, higher-dimensional problems, and robot control.

## Acknowledgments

This research has received funding from the EC's 7th Framework Programme (FP7/2007-2013), grant agreements 215078 (DIPLECS) and 247947 (GARNICS), from ELLIIT, Strategic Area for ICT research, and CADICS, funded by the Swedish Government, Swedish Research Council project ETT, and from CUAS and FOCUS, funded by the Swedish Foundation for Strategic Research.

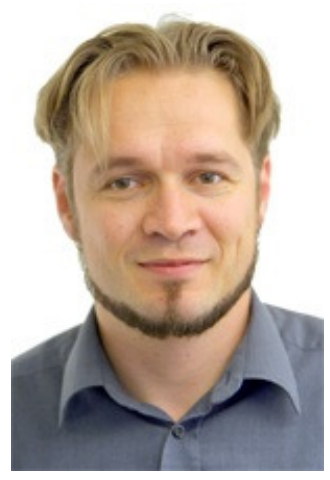

**Michael Felsberg** PhD degree (2002) in engineering from University of Kiel. Since 2008 full professor and head of CVL. Research: signal processing methods for image analysis, computer vision, and machine learning. More than 80 reviewed conference papers, journal articles, and book contributions. Awards of the German Pattern Recognition Society (DAGM) 2000, 2004, and 2005 (Olympus award), of the Swedish Society for Automated Image Analysis (SSBA) 2007 and 2010, and at Fusion 2011 (honourable mention). Coordinator of EU projects COSPAL and DIPLECS. Associate editor for the Journal of Real-Time Image Processing, area chair ICPR, and general co-chair DAGM 2011.

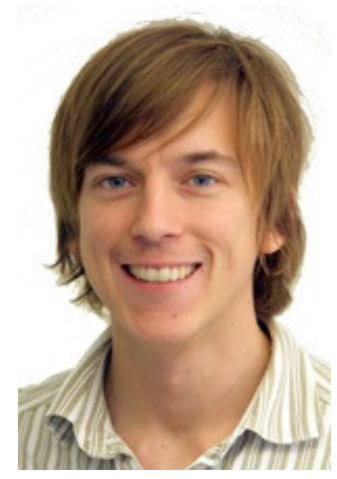

**Fredrik Larsson** Master of Science (MSc, 2007) and the PhD degree (2011) from Linköping University. From 2007 until 2011, he has been a researcher at CVL, working on the European cognitive vision projects COSPAL and DIPLECS. He has published 10 journal articles and conference papers on image processing and computer vision. Awards from the Swedish Society for Automated Image Analysis (SSBA) 2010 and at Fusion 2011 (honourable mention).

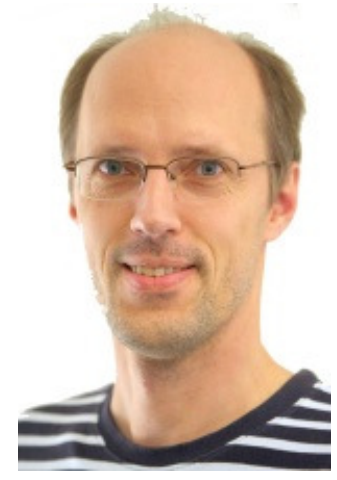

**Johan Wiklund** Master of Science (MSc, 1984) and Licentiate of Technology degree (1987) from Linköping University. Since 1984 researcher at CVL, working on image processing. Participated in various European and national projects. Responsible for system integration and demonstrator development in the European projects COSPAL, DIPLECS, and GARNICS. He has published more than 30 journal articles, book contributions and conference papers.

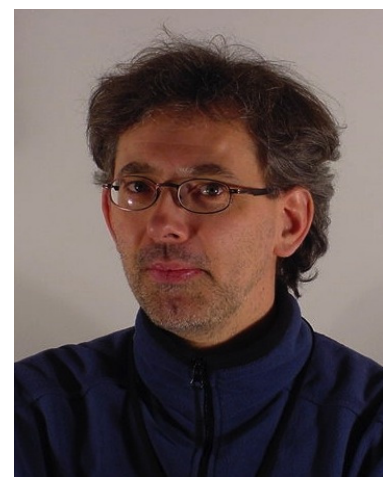

**Niclas Wadströmer** PhD from Linköping university, Sweden, 2002, in Image coding. He has been a senior lecturer in Telecommunications with the Department of Electrical Engineering, Linköping university. Currently, since 2007, he is a Scientist with the Division of Sensor and EW systems at the Swedish Defence Research agency. Among his interests are Image and signal processing, Video analytics, Multi- and hyperspectral imaging.

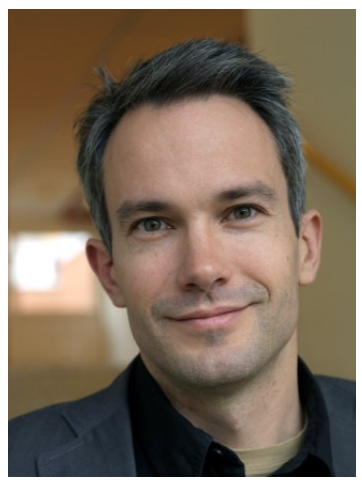

**Jörgen Ahlberg** Development Manager for Image Processing at Termisk Systemteknik AB. MSc in Computer Science and Engineering (1996) and PhD in Electrical Engineering (2002), both from Linköping University. 2002–2011 at the Swedish Defence Research Agency (FOI) as a scientist and the head of a research department. Visiting Senior Lecturer in Information Coding at Linköping University and co-founder of Visage Technologies AB. Research on image analysis, including animation and tracking of facial images, infrared and hyperspectral systems for automatic detection, recognition, and tracking, with applications as *e.g.* surveillance.